\documentclass[graybox]{svmult}

\usepackage{mathptmx}       %
\usepackage{helvet}         %
\usepackage{courier}        %
\usepackage{type1cm}        %
\usepackage{makeidx}         %
\usepackage{graphicx}        %
\usepackage{multicol}        %
\usepackage[bottom]{footmisc}%
\usepackage[linesnumbered,ruled,vlined]{algorithm2e}
\usepackage{array}
\usepackage{multirow}
\usepackage{subcaption}
\usepackage{svg}
\usepackage{color}

\usepackage{comment}
\SetKwBlock{Loop}{Loop}{EndLoop}
\usepackage{tikz}
\usetikzlibrary{shapes.multipart}
\usepackage{color}
\tikzset{
state/.style={
       rectangle split,
       rectangle split parts=4,
       rectangle split part fill={violet!30,blue!20,green!20,black!20},
       rectangle split part align={center, center, center, center},
       draw=black,%
       text width=3cm,
      inner sep=.33cm,
       text centered,
       }
}

\makeindex             %

\begin{document}

\title*{HEAL: Resilient and Self-* Hub-based Learning}
\author{Mohamed Amine Legheraba \and Stefan Galkiewicz \and Maria Potop-Butucaru \and Sébastien Tixeuil}

\authorrunning{Legheraba \and Galkiewicz \and Potop-Butucaru \and Tixeuil}

\institute{Mohamed Amine Legheraba \at Sorbonne University, CNRS, LIP6, F-75005 Paris, France, \email{mohamed.legheraba@lip6.fr}
\and Stefan Galkiewicz \at Sorbonne University, CNRS, LIP6, F-75005 Paris, France, \email{stefan.galkiewicz@lip6.fr}
\and Maria Potop-Butucaru \at Sorbonne University, CNRS, LIP6, F-75005 Paris, France, \email{maria.potop-butucaru@lip6.fr}
\and Sébastien Tixeuil \at Sorbonne University, CNRS, LIP6, F-75005 Paris, France \email{sebastien.tixeuil@lip6.fr} \and Institut Universitaire de France, Paris, France}

\maketitle

\abstract*{Decentralized learning enhances privacy, scalability, and fault tolerance by distributing data and computation across nodes. 
A popular approach is Federated learning, which relies on a central aggregator, yet faces challenges such as server vulnerabilities, scalability issues, privacy risks and most importantly, the single point of failure. Alternatively Gossip Learning and Epidemic Learning  offer fully decentralization through peer-to-peer exchanges of model updates, ensuring robustness and privacy, at the price of slower model convergence. 
In this work, we introduce a novel decentralized learning framework called HEAL. HEAL is the first cross-layer decentralized learning framework that exploits an optimized self-organizing and self-healing underlying P2P overlay combining the strengths of Federated Learning, Gossip and Epidemic Learning.
Leveraging the recently proposed Elevator algorithm, HEAL promotes dynamically chosen nodes to act as aggregators. Through simulations, we demonstrate that HEAL has similar performances to that of Federated Learning in crash-free settings, while being fully decentralized and fault-tolerant. In crash and churn prone environments HEAL outperforms Gossip and Epidemic Learning.}

\abstract{Decentralized learning enhances privacy, scalability, and fault tolerance by distributing data and computation across nodes. 
A popular approach is Federated learning, which relies on a central aggregator, yet faces challenges such as server vulnerabilities, scalability issues, privacy risks and most importantly, the single point of failure. Alternatively Gossip Learning and Epidemic Learning  offer fully decentralization through peer-to-peer exchanges of model updates, ensuring robustness and privacy, at the price of slower model convergence. 
In this work, we introduce a novel decentralized learning framework called HEAL. HEAL is the first cross-layer decentralized learning framework that exploits an optimized self-organizing and self-healing underlying P2P overlay combining the strengths of Federated Learning, Gossip and Epidemic Learning.
Leveraging the recently proposed Elevator algorithm, HEAL promotes dynamically chosen nodes to act as aggregators. Through simulations, we demonstrate that HEAL has similar performances to that of Federated Learning in crash-free settings, while being fully decentralized and fault-tolerant. In crash and churn prone environments HEAL outperforms Gossip and Epidemic Learning.}


\section{Introduction}
\label{intro}
Decentralized learning is an approach to training machine learning models, where the data and computational tasks are distributed across multiple nodes or devices, rather than centralized in a single location. 
This paradigm improves privacy, scalability, and fault tolerance by enabling participants to collaboratively train models without sharing raw data. 
Each node processes its local data and exchanges model updates (e.g., gradients or parameters) with others, using a centralized coordinator or through peer-to-peer communication. Decentralized learning is particularly beneficial in scenarios where data is naturally distributed (such as in edge computing or IoT networks), or when data privacy or resource constraints make centralization impractical. 

\textit{Related works.}
The idea of decentralized learning originates from distributed optimization and parallel computing. However, it was the advent of Federated Learning~\cite{mcmahan2017communication} that truly brought the concept into the spotlight.
In Federated Learning  all participants (nodes or devices) in a network train a machine learning model locally on their own data. Subsequently, each participant sends its model to a centralized aggregator, which aggregates the received models  and returns the global model to the nodes. This process continues until a satisfactory accuracy is achieved.
While Federated Learning provides an efficient alternative to traditional (centralized) machine learning, its dependence on a central server for aggregating models introduces several limitations. This centralization creates a \emph{single point of failure}. That is, when the server crashes the generation of the global model becomes impossible. Furthermore, the single server is an easy target for  adversarial attacks, such as model poisoning or inference attacks~\cite{rodriguez2023survey}, which can compromise the integrity of the global model. Additionally, the server's role poses scalability challenges, as it must manage potentially vast numbers of updates from distributed participants, leading to communication and computation bottlenecks. More importantly, if the server is controlled by a single organization, it could introduce biases or favor certain participants' updates, exacerbating inequalities in model performance. This phenomenon leads to \emph{learning oligarchy}.

To address these limitations, it is essential to explore totally decentralized architectures which do not rely on a central coordinator. Gossip Learning~\cite{ormandi2013gossip} and Epidemic Learning\cite{de2024epidemic} are the state of the art  approaches for fully decentralized machine learning. However, the stochastic nature of these protocols can result in slower convergence compared to Federated Learning-based approaches~\cite{hegedHus2021decentralized}. It should be noted that in the case of gossip learning, recent research has been conducted to enhance its various aspects, such as improving its security against privacy attacks~\cite{danner2015fully}, increasing its efficiency by compressing the models sent~\cite{danner2020decentralized}. More recent studies focused on examining the impact of poisoning attacks~\cite{pham2024data} on various gossip learning strategies.  

Interestingly, none of the previous decentralized federated approaches had a cross-layer design philosophy.
 
\textit{Our contribution.}
In this paper we introduce a novel form of decentralized learning, HEAL (for \textbf{H}ub-\textbf{E}nhanced \textbf{A}daptive \textbf{L}earning), which combines the benefits of Federated Learning, Gossip and Epidemic Learning (summarized in Table \ref{tab:all_algorithm}). 
HEAL leverages the network overlay created by the newly introduced Elevator algorithm~\cite{legheraba2024elevator} that promotes in a totally decentralized and adaptive manner a prescribed number of participating nodes as hubs (nodes connected to the entire network).  HEAL will use these hubs as aggregator nodes for the learning task. Unlike traditional federated learning, aggregator nodes are not pre-selected and hence HEAL becomes extremely resilient to the network dynamicity (nodes crashes and churn). That  is, HEAL self-heals and self-adapts by promoting new hubs while continuing the learning process without significant losses. We challenged HEAL, Federated Learning, Gossip Learning and Epidemic Learning with various topologies in static and dynamic environments (crash and churn prone)  on two tasks: a binary classification task using a  logistic regression~\cite{hosmer2013applied} on the Spambase dataset~\cite{spambase_94} and on a multinomial classification task using the LeNet5 model~\cite{lecun1989backpropagation} on the MNIST dataset~\cite{lecun2010mnist}.
HEAL outperforms Gossip and Epidemic Learning in both accuracy and convergence time. Moreover, HEAL is resilient to churn and crashes while Federated Learning cannot cope with these faults. 

\begin{table}
\caption{Comparison of different decentralized learning algorithms.}
\label{tab:all_algorithm}
\begin{center}
\begin{tabular}{|m{3cm}|m{3cm}|c|m{1.5cm}|m{2cm}|}
\hline
\textbf{Decentralized Federated Learning} & \textbf{Topology} & \textbf{Fault resilience} & \textbf{Churn resilience} & \textbf{Aggregation speed} \\
\hline
Federated Learning & Static (Star~\cite{mcmahan2017communication} and Multi-Star~\cite{hsieh2017gaia}) & No & No & quick at the server \\
\hline
\multirow{2}{*}{Gossip Learning} & Static (Random Regular~\cite{hegedHus2021decentralized}) & No & No & slow local \\ \cline{2-5}
 & Dynamic (with Newscast~\cite{ormandi2013gossip}) & Yes & Yes & slow local \\
\hline
\multirow{2}{*}{Epidemic Learning} & Dynamic (Newscast~\cite{de2024epidemic}) & Yes & Yes & slow local \\ \cline{2-5}
 & Dynamic (FedLay~\cite{hua2024towards}) & Yes (only one) & No & slow local \\
\hline
\textbf{HEAL (this paper)} & \textbf{Dynamic (Elevator~\cite{legheraba2024elevator})} & \textbf{Yes} & \textbf{Yes} & \textbf{quick at the hubs} \\
\hline
\end{tabular}
\end{center}
\end{table}
\vspace{-1cm}
\paragraph{Paper organization}
In Section \ref{sec:learning background} we propose an overview of decentralized learning strategies. Section \ref{sec:HEAL} introduces the architecture of HEAL and the detailed description of the HEAL learning strategies. Section \ref{sec:evaluation} presents our extensive evaluations. Section \ref{sec:conclusions} concludes and proposes open research directions.
\section{Overview of Decentralized Learning Techniques}
\label{sec:learning background}
The distinguishing factors among various decentralized learning algorithms in the literature are: (1) the algorithm employed to propagate and aggregate  learning models within the network, and (2) the network topology on which the learning occurs. Naturally, these two concepts are interconnected, as certain propagation methods are better suited to specific topologies. 

\paragraph{Decentralized propagation and aggregation of the learning models.}

There are various techniques for propagating the learning models within a network, each with its own set of advantages and disadvantages. The three most commonly discussed methods in the literature are \emph{Federated Learning}, \emph{Gossip Learning}, and \emph{Epidemic Learning}. In Federated Learning~\cite{mcmahan2017communication}, all models are aggregated at a central server, facilitating rapid convergence towards a global model. In Gossip Learning~\cite{ormandi2013gossip}, each participant shares its model at a specified time interval with a randomly chosen neighbor in the network. In Epidemic Learning~\cite{de2024epidemic}, each participant shares its model with all their neighbors in each cycle.
 
\paragraph{Network topology vs decentralized learning.}
In decentralized learning, the network topology significantly influences the performance of the learning task, as evidenced by various studies in the literature~\cite{vogels2022beyond}. Different topologies impact the speed of information propagation and the system's resilience to node or link failures. At the extremes, we have the star topology, typically used with Federated Learning, and the random graph topology, often paired with Gossip Learning. However, many other topologies exist between these extremes. Additionally, it is important to distinguish between static topologies (predefined and unchangeable) and dynamic topologies (which evolve over time).

\begin{itemize}
\item{Static topologies.} Among static topologies,  \emph{star topology}, features a central server and clients connected to it. This setup is not entirely decentralized, as the aggregator server is selected at the beginning of the learning process, creating a single point of failure. The \emph{multi-star topology} is a variation of the star topology, involving multiple servers. Typically, all stars are interconnected in a complete topology, with all other nodes connected to a predefined star. Next, we have the \emph{complete topology}, where every participant communicates directly with all others. While this maximizes information propagation speed and ensures rapid convergence, it is impractical for large-scale networks. Another common topology is the \emph{ring topology}, where each node is connected to two neighbors, forming a circular ring. This topology is straightforward to construct but does not scale well. Finally,  in \emph{random regular graphs} each participant is randomly connected to \emph{k} other participants in the network, with \emph{k} being a predefined parameter. This topology is highly robust against failures and churn, but the learning process convergence is slow. Other random topologies in the literature include small worlds and power-law networks.

\item{Self-organised dynamic topologies.}
To establish a dynamic topology, two primary approaches are discussed in the literature: Distributed Hash Table (DHT)-based methods and decentralized peer sampling methods. 
Among the DHT-based methods, Chord~\cite{stoica2001chord} is a notable example. Additionally, Fedlay~\cite{hua2024towards} is specifically designed for decentralized learning. For non-DHT methods, Newscast~\cite{jelasity2007gossip} is commonly used in Gossip Learning and Epidemic Learning.
\end{itemize}
An important aspect to consider in dynamic topologies is their resilience to failures. A specific type of failure is churn, where nodes in a peer-to-peer network enter and leave without any control. Another particular case of failures involves attacks targeting specific nodes, such as servers or central nodes. Centralized topologies are highly sensitive to these types of failures.
In a static topology, there is no possibility to repair the topology in the event of a failure. In a DHT, some repairs are possible, but not always guaranteed face to high churn. 
Peer sampling algorithms are a good compromise to repair the topology when failures are detected and to be resilient to high churn. 

HEAL, uses as underlying topology Elevator~\cite{legheraba2024elevator}, a recently proposed decentralized peer-sampling algorithm. This algorithm enables nodes in a peer-to-peer network to construct an overlay with \emph{h} defined hubs, each hub being connected to all nodes in the network, with \emph{h} being a parameter of the algorithm. Elevator is totally distributed, self-organizing and resilient to churn.
In Table~\ref{tab:all_algorithm}, we summarized the topologies used with various decentralized learning algorithms, along with the strengths and weaknesses of each approach.

\begin{figure}
\caption{HEAL architecture}
\label{heal:architecture}
\begin{center}
 \begin{tikzpicture}
  \node [state] (box) at (2,0) {
            Supervised Learning Task (Ex: LeNet5)
            \nodepart{two}
            HEAL-Learning Protocol
            \nodepart{three}
            HEAL-Overlay (Elevator)
            \nodepart{four}
            Network Layer (e.g. TCP/IP)
        };

        \foreach \n/\l in {
            text/Application Layer, 
            two/Model Aggregation Layer, 
            three/Peer-Sampling Layer,
            four/Physical and logical Network Layer
        } {
            \draw[->] ([xshift=-0.5cm]box.\n\space east) -- ++(2,0) node[right] {\l};
        }
\end{tikzpicture}
\end{center}
\end{figure}
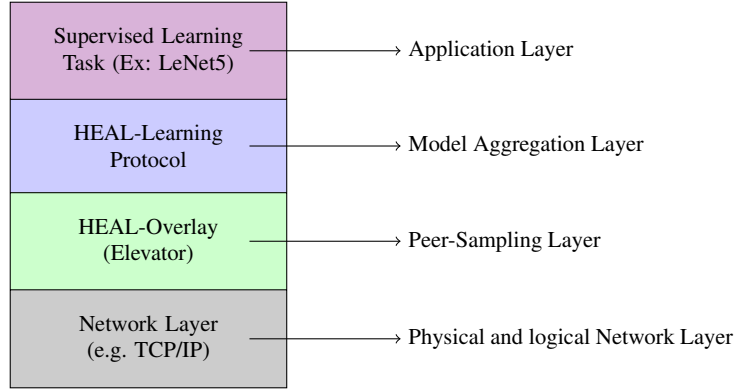

\vspace{-1cm}
\section{HEAL architecture}
\label{sec:HEAL}
Federated Learning is vulnerable due to its reliance on a central server, so our protocol must avoid having a single point of failure. Additionally, to ensure resilience to failures and churn, the topology should not be predefined but generated in a peer-to-peer manner. Conversely, to guarantee rapid model convergence, learning models should not be shared via gossip within the network but aggregated by a network node, which will then create the global model and distribute it back to the other network nodes. These two aspects may seem contradictory, but the HEAL-overlay (Elevator) protocol allows us to create a topology that satisfies both requirements. 

HEAL architecture shown in Figure \ref{heal:architecture} is composed of two layers on top of the physical network. HEAL-overlay give by the Elevator protocol introduced in ~\cite{legheraba2024elevator} and HEAL-learning protocol described in the sequel.

\begin{algorithm}[htbp]
  \caption{HEAL Learning: The Hub algorithm}
  \label{algo:HubLearningHub}
  \KwData{duration to wait for model: $\mathit{delta\_time}$}
  \KwData{list of all hubs (obtained by the HEAL overlay, Elevator protocol): $\mathit{hubs\_list}$}
  $\mathit{nb\_hubs} \gets \mathit{hubs\_list.size}()$\\
  \Loop{ 
    $\mathit{models} \gets \{\}$ \\
    $\mathit{time} \gets \mathit{time.now}()$ \\
    $\mathit{backwards\_list} \gets \{\}$ \\
    \While{$\mathit{time.now}() < \mathit{time}+\mathit{delta\_time}$}{
      $\mathit{peer\_model}, \mathit{peer} \gets \mathit{receive()}$ \\
      $\mathit{models.append}(\mathit{peer\_model})$ \\
      $\mathit{backwards\_list.append(peer)}$
    }
    $\mathit{average\_model} \gets \mathit{average}(\mathit{models})$ \\
    $\mathit{send}(\mathit{hubs\_list}, \mathit{average\_model})$ \\
    $\mathit{models\_hubs} \gets \{\}$ \\
    $\mathit{models\_hubs.append}(\mathit{average\_model})$ \\
    $\mathit{nb\_receive} \gets 0$\\
    \While{$\mathit{nb\_receive} < \mathit{nb\_hubs - 1}$}{
      $\mathit{hub\_model} \gets \mathit{receive}()$ \\
      $\mathit{nb\_receive}++$ \\
      $\mathit{models\_hubs.append}(\mathit{hub\_model})$
    }
    $\mathit{global\_model} \gets \mathit{average}(\mathit{models\_hubs})$ \\
    $\mathit{send}(\mathit{backwards\_list}, \mathit{global\_model})$ 
  }
\end{algorithm}

\begin{algorithm}[htbp]
  \caption{HEAL Learning: The client algorithm}
  \label{algo:HubLearningNode}
  \KwData{duration to wait for model: $\mathit{delta\_time}$}
  \KwData{The model, same for all node, weights or parameters initialized at random: $\mathit{model}$}
  \KwData{The local data of the node): $\mathit{data}$}
  \KwData{list of all hubs (obtained by HEAL overlay, Elevator protocol): $\mathit{hubs\_list}$}
  \KwData{number of hub to send the model: $\mathit{number\_hub\_send}$}  %
  \Loop{ 
    $\mathit{model} \gets \mathit{trainModel}(\mathit{model}, \mathit{data})$ \\
    $\mathit{hubs} \gets \mathit{chooseRandom}(\mathit{hubs\_list}, \mathit{number\_hub\_send})$ \\
    \For{$\mathit{hub} \in \mathit{hubs}$}{
        $\mathit{send}(\mathit{hub}, \mathit{model})$ \\
    }
    $\mathit{hubs\_models} \gets \mathit{list}()$ \\
    \tcp{Receiving the global models from the hubs}
    \For{$\mathit{hub} \in \mathit{hubs}$}{
        $\mathit{model\_hub} \gets \mathit{receive}()$ \\
        $\mathit{hubs\_models.append}(\mathit{model\_hub})$ \\
    } 
    $\mathit{model} \gets \mathit{average}(\mathit{hubs\_models})$
  }
\end{algorithm}

\subsection{HEAL Overlay}
In the following we briefly revisit how Elevator operates. For a more detailed explanation, readers can refer to the article that introduces the algorithm~\cite{legheraba2024elevator}.
 
The Elevator protocol performs the following actions during each cycle: Each node in the peer-to-peer network retrieves the list of neighbors of their neighbors (i.e., the neighbors at a distance of two). The node then constructs an ordered list of the most frequent peers (the frequency map) and contacts the \emph{c} most frequent nodes (referred to as \emph{preferred}). Each contacted node responds by sending the addresses from its backward list to the contacting node and adds the contacting node to its backward list. The contacting node's cache is then reset to an empty array. Subsequently, the node selects the \emph{h} most frequent peers and \emph{c-h} random peers from the backward lists of all preferred peers to populate its cache.

This protocol enables the rapid formation (in 4 cycles or fewer in practical settings) of a network topology with \emph{h} defined hubs and a random distribution of the remaining incoming connections. The resulting network has a diameter of 2 and is highly resistant to both failures (including hub failures) and churn. In the event of all hubs failing, new hubs quickly emerge (typically  with one cycle). These properties are particularly advantageous for decentralized learning, suggesting that we can implement a learning algorithm on this topology that achieves performance levels similar to Federated Learning while maintaining resilience properties as Gossip and  Epidemic Learning.
 
\subsection{HEAL  Learning Protocol}
Regarding the communication algorithm, we utilize the hubs within the network as aggregators, similar to how the central server aggregates models in Federated Learning. The key difference is that multiple hubs perform the aggregation, not just one, and these hubs emerge automatically. We leverage the presence of multiple hubs to distribute the aggregation workload, with each hub handling a portion of the network nodes and subsequently aggregating with each other. Each hub then returns the global model to its clients, and the protocol begins a new cycle. As with Federated Learning, the learning process continues over several cycles and concludes when the global model has converged.

In the event of one or more hubs failing during a protocol cycle, the remaining hubs can temporarily manage the nodes without a dedicated hub until a new hub emerges, which typically occurs within two cycles. If all hubs fail, the aggregation process halts but resumes as soon as new hubs appear, again within two cycles. It's important to note that Elevator also establishes random connections in the network, in addition to hub connections. In HEAL, we focus solely on connections to hubs (and between hubs) for model aggregation. These random connections could potentially be used to accelerate model convergence or mitigate malicious behavior, but this is beyond the scope of our current work. For now, we assume that all network nodes (and hubs) are honest, with plans to investigate malicious behavior in future research.

One intriguing feature of Elevator is the ability to select the number of hubs in the network through a parameter shared by all nodes. This is particularly valuable in HEAL, as it allows us to balance between having fewer hubs for higher convergence speed and more hubs for greater resilience to failures. Additionally, HEAL offers the flexibility to choose the number of hubs to which a client sends its model. A more detailed description of how   Learning operates in HEAL follows.

Each node in the network executes the HEAL learning protocol, in addition to HEAL overlay construction via the Elevator protocol (that dynamically assigns "normal" (client) or "hub"(server) status to the participating nodes):

\begin{itemize} 
\item If the node is a normal(client) node, it 
    (1) selects a number of hubs(servers) at random, 
    (2) performs a local training step, 
    (3) sends the trained model to the hubs, 
    and (4) waits for the global model. 
\item If the node is a hub, it 
    (1) waits for a \emph{delta} period to receive models from normal nodes, 
    (2) aggregates these models by averaging their parameters, 
    and (3) sends its aggregated model to all other hubs. 
    (4) It then waits to receive models from other hubs and 
    (5) aggregates all these models to obtain the global model. 
    (6) Finally, the hub sends the global model back to the nodes.
\end{itemize}

 Algorithms~\ref{algo:HubLearningHub} and \ref{algo:HubLearningNode} present the detailed pseudo-code of HEAL Learning.

\section{Evaluation results}
\label{sec:evaluation}
We evaluated our algorithm using simulations on the Gossipy simulator~\footnote{https://github.com/makgyver/gossipy}. We compared Hub Learning against Federated Learning~\cite{mcmahan2017communication}, Gaia~\cite{hsieh2017gaia}, Gossip Learning~\cite{ormandi2013gossip}, Epidemic Learning~\cite{de2024epidemic}, Epidemic Learning on a Chord topology~\cite{stoica2001chord}, Epidemic Learning on a ring topology, and Fedlay~\cite{hua2024towards}. For the static topologies (Federated Learning, ring, Chord, and Gaia), we generated the topology using the Python library Networkx~\footnote{https://networkx.org/}. 
For the dynamic topologies (Gossip Learning, Epidemic Learning, Fedlay and Hub Learning with Elevator), we generated the topology using the PeerSim simulator~\cite{p2p09-peersim}. 
In Elevator (used by Hub Learning), the connections are directional. However, to compare them with other algorithms (which assume an undirected graph), we modified the underlying graph of the topology generated to make it undirected. 
All evaluations were conducted with a network of 100 nodes. For Elevator, we used 5 hubs, as we found this number to be a good balance between performance and resilience. For Gaia, we had 5 servers responsible for aggregation (to compare with the 5 hubs) and 19 nodes (or workers) attached to each server. Each algorithm was evaluated 5 times, and we present the average results obtained.

We assessed all protocols on two tasks: a binary classification task (Logistic Regression~\cite{hosmer2013applied} on the Spambase dataset~\cite{spambase_94}, with a learning rate of 0.1) and on a multinomial classification task (LeNet5~\cite{lecun1989backpropagation} on the MNIST dataset~\cite{lecun2010mnist}, with a learning rate of 0.001). The weight decay (regularization  parameter) was fixed at 0.01. Our algorithm was evaluated under various conditions: the failure of 20\% of nodes, the failure of a hub, the failure of all 5 hubs, and during churn (where 10\% of nodes disappear at each cycle and are replaced by new nodes).

All simulations were run on 16 vCPU, using 64G of memory, on a cluster composed of 10 servers.

\begin{figure}
\caption{Accuracy of various communication protocols, for the MNIST dataset, with a network of 100 nodes, during 1000 cycles. HEAL overlay has 5 hubs, each node sends its model to one hub.}
\begin{center}
\begin{subfigure}{0.4\textwidth}
    \caption{Without failures}
    \label{fig:AccuracyMNIST}    \includegraphics[width=\textwidth]
    {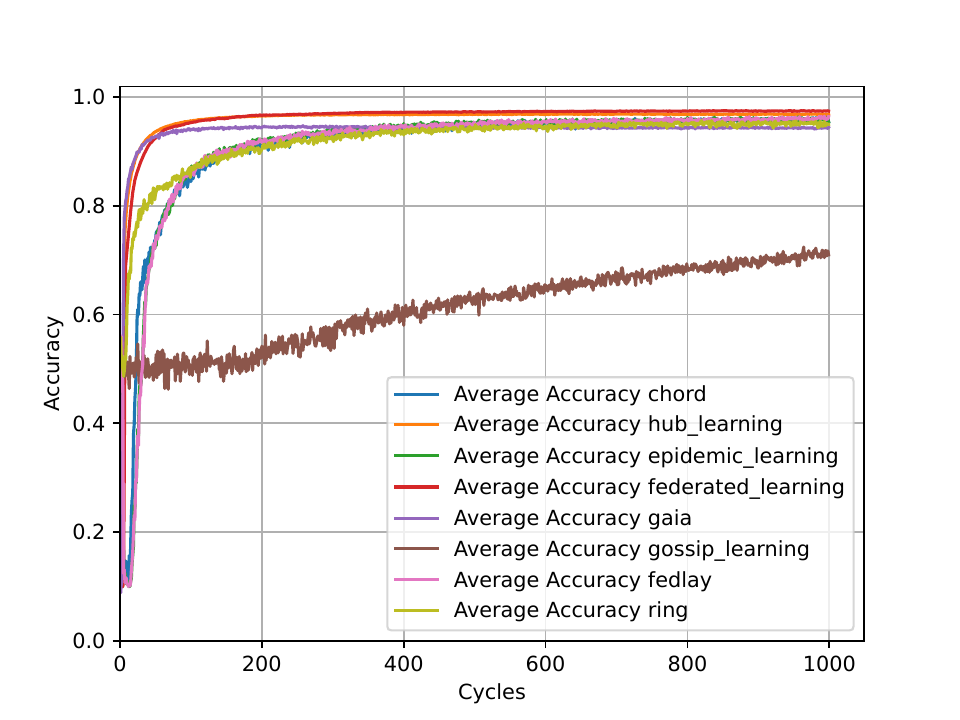}
\end{subfigure}
\begin{subfigure}{0.4\textwidth}
    \caption{When 20\% of the nodes fail at round 10}
    \label{fig:AccuracyCrash20peers}
    \includegraphics[width=\textwidth]
    {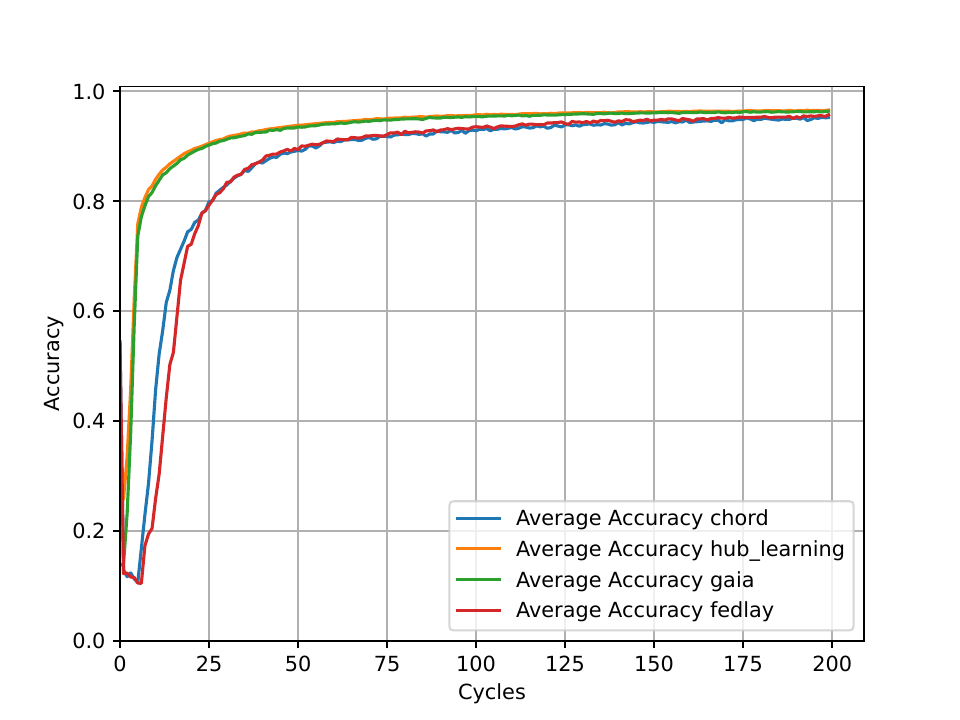}
\end{subfigure}
\hfill
\end{center}
\end{figure}

\textit{Crash-free, churn-free environment:}
For simulations without failures and churn, we ran all algorithms over 1000 cycles, on the two learning tasks (Spambase, MNIST). As shown in the Figure~\ref{fig:AccuracyMNIST}, when there are no failures in the network, Federated Learning performs best, which was expected. Surprisingly, the ring topology performs second best despite lower connectivity. 
Other topologies based on random graphs perform less well. HEAL, however, performs very well for both the Spambase and MNIST datasets.

In Tables~\ref{tab:all_results} we have compiled the results for all learning algorithms, for the two learning tasks, with the final accuracy, obtained after 1000 cycles. Federated Learning, Gaia, and HEAL have very similar results, with a final accuracy of around 0.88 on Spambase, and 0.95 on MNIST. Gossip-based approaches achieve values of 0.85 and 0.91 on Spambase and MNIST. 
Convergence time is very fast for aggregator-based approaches: on Spambase, Federated Learning converges to 0.85 in 2 cycles, and HEAL takes 4 cycles. On the other hand, to converge on 0.90 accuracy, HEAL takes 339 cycles while Federated Learning takes 135 cycles, hinting at possible HEAL optimization, e.g. adjusting the learning rate. 
On MNIST, HEAL performs very well, even better than Federated Learning, and it converges to 0.95 accuracy in 76 cycles.%

\begin{figure}
\caption{Accuracy of HEAL for the MNIST dataset, with 100 nodes}
\begin{subfigure}{0.48\textwidth}
    \caption{HEAL with different numbers of hubs (\emph{h}), from 1 to 25, each node sent its model to (\emph{s}) hubs, with (\emph{s}) equals to 1 or $\frac{h}{2}$, no failures, 2000 cycles}
    \label{fig:AccuracyVariousNbHubs}
    \includegraphics[width=\textwidth]
    {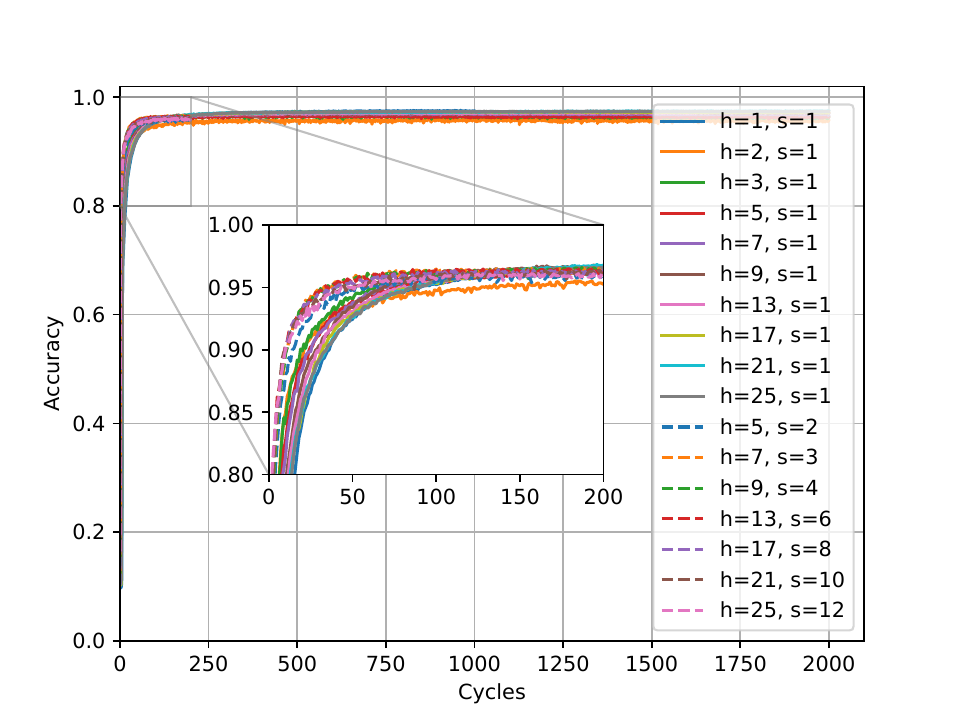}
\end{subfigure}
\hfill 
\begin{subfigure}{0.48\textwidth}
    \caption{HEAL for all contexts (no failures, crash of 20 peers, crash of 1 hub, crash of all hubs, churn), with 5 hubs, each node sent its model to one hub, 200 cycles}
    \label{fig:AccuracyContexts}    \includegraphics[width=\textwidth]
    {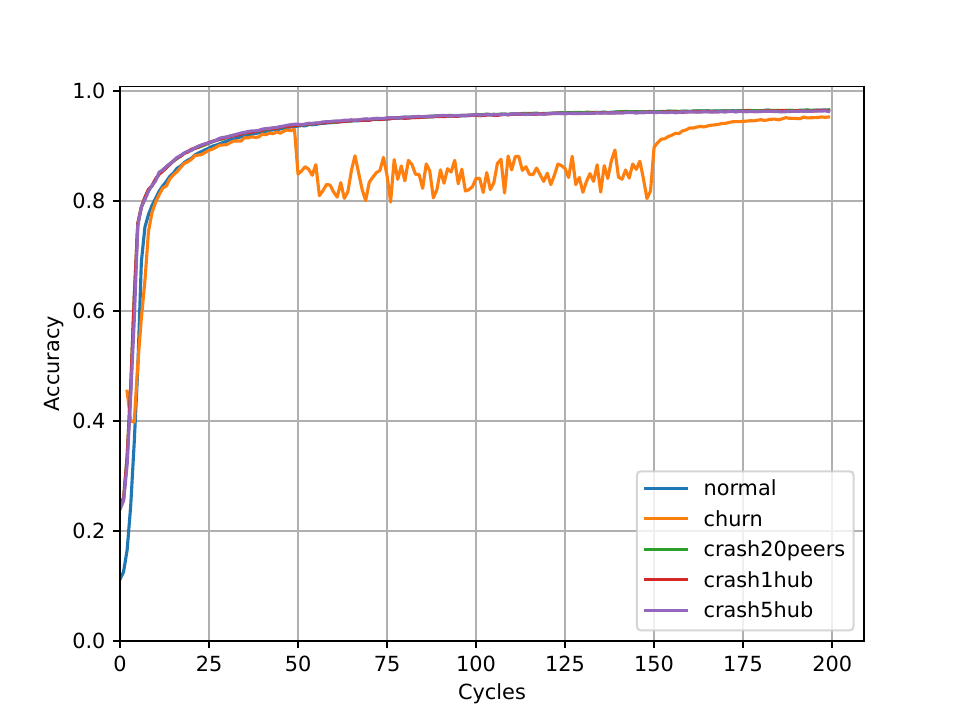}
\end{subfigure}
\end{figure}

\textit{HEAL parameterized with number of hubs and number chosen hubs:}
We ran simulations of HEAL, changing the number of hubs over 2000 cycles. On Figure~\ref{fig:AccuracyVariousNbHubs}, we observe that increasing the number of hubs (and the number of hubs to which clients send their model) has almost no impact on accuracy, which is expected since hubs aggregate models. There is a slight drop in accuracy when the number of hubs is increased significantly, due to the fact that only non-hubs are learning, not hubs. Increasing the number of hubs to which we send our model, from 1 to \emph{$nb\_hubs/2$}, slightly increases convergence speed.

\textit{Crashes-prone environment:}
We analyze the performance of the algorithms when
the network suffers crashes. To simulate a brutal failure we
disconnected 20\% of the nodes chosen uniformly at random, just after the start of the learning process, i.e., in this case, we have disconnected 20 nodes at cycle 10, and we compared HEAL with Chord, Gaia and Fedlay.
HEAL is the algorithm with the best results, although Gaia remains very close, as seen in Figure~\ref{fig:AccuracyCrash20peers}.

\textit{HEAL under churn environment and hub-targeted attacks:}
We further analyzed the performance of HEAL under network churn conditions. To simulate churn, we disconnected 10\% of the nodes at each cycle and replaced them with an equal number of new nodes, each connected to 20 nodes uniformly at random, between cycles 50 and 150.
We also analyzed the performance of the main learning algorithms after a targeted attack on the hubs during the execution of the simulation. We tested two scenarios, one where we disconnected one of the 5 hubs, and another where we disconnected all 5 hubs at the same time. In both case the failure happened in the round 10. In Figure~\ref{fig:AccuracyContexts}, we have compared the execution of HEAL without failures, and with different failure scenarios (crash of 20 nodes, crash of one hub, crash of all 5 hubs, churn). As can be seen, there is no significant impact when 20 nodes or hubs crash. Indeed, thanks to the Elevator overlay, even when all the hubs are shutdown, 5 new nodes are elected very quickly as hubs, and the training job continues as if no catastrophic event had happened. During churn, model accuracy falls slightly, but rises again very quickly once churn is over, back to the level without failures.

\section{Conclusion}
\label{sec:conclusions}
In this paper we  introduced HEAL protocol for decentralized learning that combines the convergence speed of Federated Learning with the resilience to churn and failures of Gossip and Epidemic Learning. Our simulation results (summarized in Table~\ref{tab:all_results}) show that, on the MNIST dataset, HEAL (with 5 hubs) achieves an accuracy 136\% higher than Gossip Learning, 106\% higher than Epidemic Learning and 99\% of the accuracy of the baseline Federated Learning. HEAL  achieves an accuracy of 0.95 in 76 cycles, which is one cycle slower than Gaia, and much faster than random graph methods, which achieve this value in 5 times as many cycles. By setting HEAL with 7 hubs and the number of hubs to which each node sends its model at 3, it is possible to reduce it to 33 cycles, which is 2.3 times faster than Gaia (the second best result). Our protocol continues to operate in the presence of faults, and in each fault scenario, the final accuracy is at most equal to 98\% of the accuracy in a fault-free context. HEAL paves the way for a new approach to decentralized learning, featuring a cross-layer approach. Our future work will focus on adapting HEAL to heterogeneous environments,  enhancing its robustness against various attacks (e.g. poisoning attacks, model attacks, etc). %
\begin{acknowledgement}
The work presented in this document has received funding from the EU Horizon Europe research and innovation Programme under Grant Agreement No. 101070118.
\end{acknowledgement}

\begin{table}
\caption{Final accuracy by communication method and dataset used. HEAL  overlay with 5 hubs, each node sends its model to one hub (fault and churn free scenario).}
\label{tab:all_results}
\centering
\begin{tabular}{|c|c|c|}
\hline
\textbf{Method} & \textbf{Spambase} & \textbf{MNIST (LeNet)} \\
\hline
Federated Learning & 0.9087 & 0.9742 \\
\hline
Gaia & 0.8826 & 0.9442 \\
\hline
Gossip Learning & 0.8322 & 0.7098 \\
\hline
Epidemic Learning & 0.9548 & 0.9111 \\
\hline
Ring & 0.9076 & 0.9499 \\
\hline
Chord & 0.9063 & 0.9542 \\
\hline
FedLay & 0.9056 & 0.9639 \\
\hline
HEAL & 0.9001 & 0.9687 \\
\hline
\end{tabular}
\end{table}

\bibliographystyle{spmpsci}
\bibliography{ref}

\begin{thebibliography}{10}
\providecommand{\url}[1]{{#1}}
\providecommand{\urlprefix}{URL }
\expandafter\ifx\csname urlstyle\endcsname\relax
  \providecommand{\doi}[1]{DOI~\discretionary{}{}{}#1}\else
  \providecommand{\doi}{DOI~\discretionary{}{}{}\begingroup
  \urlstyle{rm}\Url}\fi

\bibitem{danner2020decentralized}
Danner, G., Heged{\H{u}}s, I., Jelasity, M.: Decentralized machine learning
  using compressed push-pull averaging.
\newblock In: Proceedings of the 1st International Workshop on Distributed
  Infrastructure for Common Good, pp. 31--36 (2020)

\bibitem{danner2015fully}
Danner, G., Jelasity, M.: Fully distributed privacy preserving mini-batch
  gradient descent learning.
\newblock In: Distributed Applications and Interoperable Systems: 15th IFIP WG
  6.1 International Conference, DAIS 2015, Held as Part of the 10th
  International Federated Conference on Distributed Computing Techniques,
  DisCoTec 2015, Grenoble, France, June 2-4, 2015, Proceedings 15, pp. 30--44.
  Springer (2015)

\bibitem{de2024epidemic}
De~Vos, M., Farhadkhani, S., Guerraoui, R., Kermarrec, A.M., Pires, R., Sharma,
  R.: Epidemic learning: Boosting decentralized learning with randomized
  communication.
\newblock Advances in Neural Information Processing Systems \textbf{36} (2024)

\bibitem{hegedHus2021decentralized}
Heged{\H{u}}s, I., Danner, G., Jelasity, M.: Decentralized learning works: An
  empirical comparison of gossip learning and federated learning.
\newblock Journal of Parallel and Distributed Computing \textbf{148}, 109--124
  (2021)

\bibitem{spambase_94}
Hopkins~Mark Reeber~Erik, F.G., Jaap, S.: {Spambase}.
\newblock UCI Machine Learning Repository (1999).
\newblock {DOI}: https://doi.org/10.24432/C53G6X

\bibitem{hosmer2013applied}
Hosmer~Jr, D.W., Lemeshow, S., Sturdivant, R.X.: Applied logistic regression.
\newblock John Wiley \& Sons (2013)

\bibitem{hsieh2017gaia}
Hsieh, K., Harlap, A., Vijaykumar, N., Konomis, D., Ganger, G.R., Gibbons,
  P.B., Mutlu, O.: Gaia:$\{$Geo-Distributed$\}$ machine learning approaching
  $\{$LAN$\}$ speeds.
\newblock In: 14th USENIX Symposium on Networked Systems Design and
  Implementation (NSDI 17), pp. 629--647 (2017)

\bibitem{hua2024towards}
Hua, Y., Pang, J., Zhang, X., Liu, Y., Shi, X., Wang, B., Liu, Y., Qian, C.:
  Towards practical overlay networks for decentralized federated learning.
\newblock arXiv preprint arXiv:2409.05331  (2024)

\bibitem{jelasity2007gossip}
Jelasity, M., Voulgaris, S., Guerraoui, R., Kermarrec, A.M., Van~Steen, M.:
  Gossip-based peer sampling.
\newblock ACM Transactions on Computer Systems (TOCS) \textbf{25}(3), 8--es
  (2007)

\bibitem{lecun1989backpropagation}
LeCun, Y., Boser, B., Denker, J.S., Henderson, D., Howard, R.E., Hubbard, W.,
  Jackel, L.D.: Backpropagation applied to handwritten zip code recognition.
\newblock Neural computation \textbf{1}(4), 541--551 (1989)

\bibitem{legheraba2024elevator}
Legheraba, M.A., Potop-Butucaru, M., Tixeuil, S.: Elevator: Self-* and
  persistent hub sampling service in unstructured peer-to-peer networks.
\newblock arXiv preprint arXiv:2406.07946  (2024)

\bibitem{mcmahan2017communication}
McMahan, B., Moore, E., Ramage, D., Hampson, S., y~Arcas, B.A.:
  Communication-efficient learning of deep networks from decentralized data.
\newblock In: Artificial intelligence and statistics, pp. 1273--1282. PMLR
  (2017)

\bibitem{p2p09-peersim}
Montresor, A., Jelasity, M.: {PeerSim}: A scalable {P2P} simulator.
\newblock In: Proc. of the 9th Int. Conference on Peer-to-Peer (P2P'09), pp.
  99--100. Seattle, WA (2009)

\bibitem{ormandi2013gossip}
Orm{\'a}ndi, R., Heged{\H{u}}s, I., Jelasity, M.: Gossip learning with linear
  models on fully distributed data.
\newblock Concurrency and Computation: Practice and Experience \textbf{25}(4),
  556--571 (2013)

\bibitem{pham2024data}
Pham, A., Potop-Butucaru, M., Tixeuil, S., Fdida, S.: Data poisoning attacks in
  gossip learning.
\newblock In: International Conference on Advanced Information Networking and
  Applications, pp. 213--224. Springer (2024)

\bibitem{rodriguez2023survey}
Rodr{\'\i}guez-Barroso, N., Jim{\'e}nez-L{\'o}pez, D., Luz{\'o}n, M.V.,
  Herrera, F., Mart{\'\i}nez-C{\'a}mara, E.: Survey on federated learning
  threats: Concepts, taxonomy on attacks and defences, experimental study and
  challenges.
\newblock Information Fusion \textbf{90}, 148--173 (2023)

\bibitem{stoica2001chord}
Stoica, I., Morris, R., Karger, D., Kaashoek, M.F., Balakrishnan, H.: Chord: A
  scalable peer-to-peer lookup service for internet applications.
\newblock ACM SIGCOMM computer communication review \textbf{31}(4), 149--160
  (2001)

\bibitem{vogels2022beyond}
Vogels, T., Hendrikx, H., Jaggi, M.: Beyond spectral gap: The role of the
  topology in decentralized learning.
\newblock Advances in Neural Information Processing Systems \textbf{35},
  15,039--15,050 (2022)

\bibitem{lecun2010mnist}
Yann~LeCun, C.C., Burges, C.J.: Mnist database of handwritten digits (2010).
\newblock \urlprefix\url{https://yann.lecun.com/exdb/mnist/}.
\newblock Accessed: 2024-12-13

\end{thebibliography}

\end{document}